\documentclass[10pt,twocolumn,letterpaper]{article}

\usepackage[pagenumbers]{cvpr}

\usepackage{cuted}

\usepackage{currfile}

\usepackage{caption}

\usepackage{xspace}
\newcommand{\modelname}{CoSMo\xspace}
\newcommand{\modelfullname}{CoSMo: A Compact Spectral Space of Body Motion in Video}
\definecolor{darkgreen}{rgb}{0.1, 0.4, 0.0}
\definecolor{darkgrey}{rgb}{0.4, 0.3, 0.4}

\AtBeginDocument{%
  \setlength{\abovedisplayskip}{4pt plus 1pt minus 1pt}%
  \setlength{\belowdisplayskip}{4pt plus 1pt minus 1pt}%
  \setlength{\abovedisplayshortskip}{1pt plus 1pt}%
  \setlength{\belowdisplayshortskip}{3pt plus 1pt minus 1pt}}

\usepackage{animate}
\usepackage[pagebackref,breaklinks,colorlinks,citecolor=Green,bookmarks=false]{hyperref}

\def\paperID{7232}
\def\confName{CVPR}
\def\confYear{2027}

\title{\modelfullname}
\author{Maham Tanveer \qquad Hao Zhang\\
Simon Fraser University}
\date{September 2026}

\begin{document}
\raggedbottom

\twocolumn[{%
\renewcommand\twocolumn[1][]{#1}%
\maketitle
\begin{center}
  \vspace{-14pt}%
  \centering
  \captionsetup{type=figure,skip=4pt}
  \input{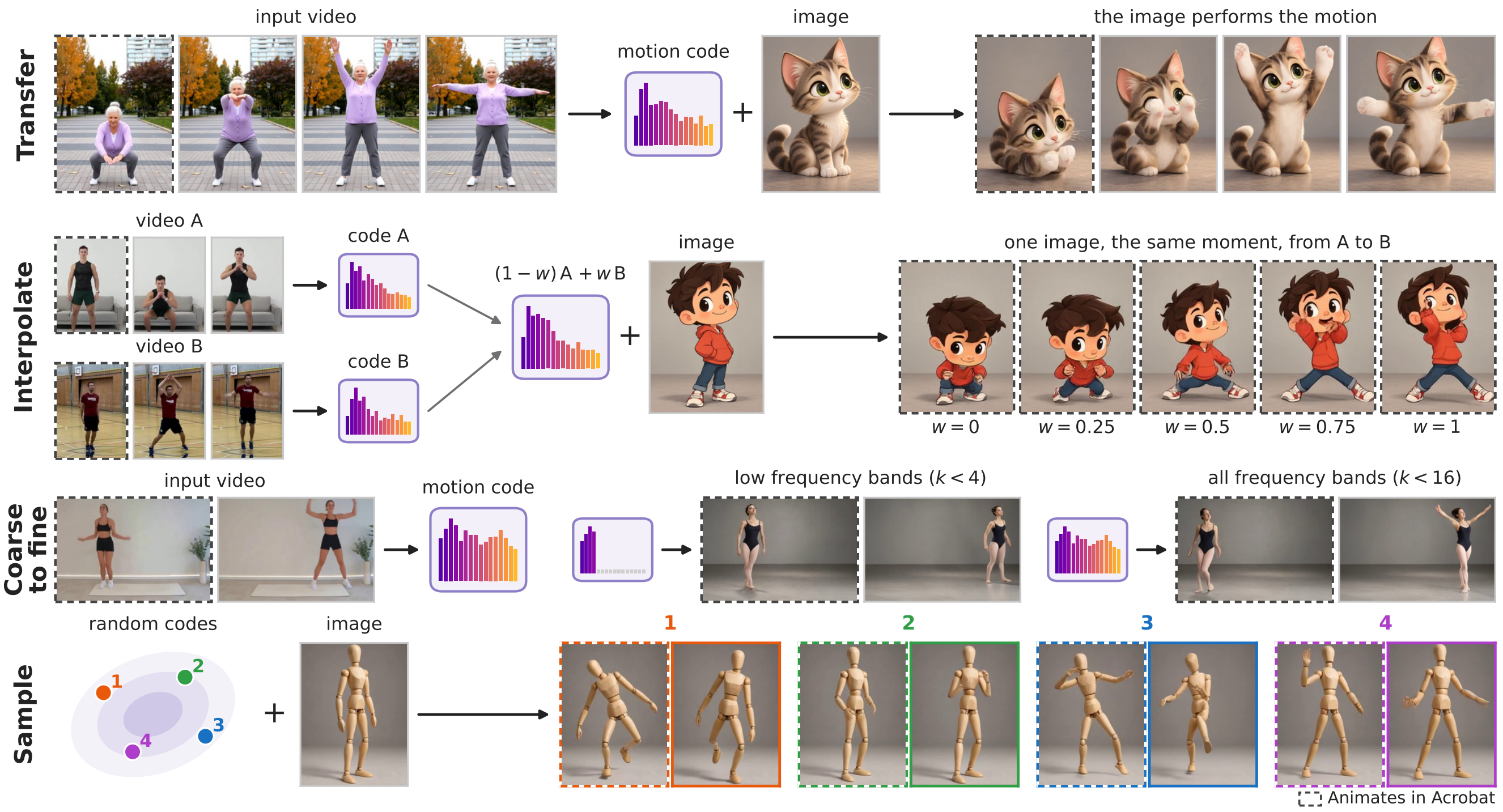}%
  \captionof{figure}{
  Our spectral space of body motion in video, with \textbf{one compact code per motion}, enables several applications.  \emph{Transfer:} the code of the input video
  makes a still image perform the same motion, on a body with very different proportions. \emph{Interpolate:} blending the codes of a squat (A) and a jumping jack
  (B) with weight $w$ turns one motion gradually into the other. \emph{Coarse to fine:} keeping only the 4 lowest of the code's 16 bands (left) gives the large
  left-to-right motion, and the full code (right) adds the finer arm details. \emph{Sample:} random codes drawn from the space are applied to the same image, with
  no source video. Dashed-box frames animate in Adobe Acrobat.
  }
  \label{fig:teaser}
  \vspace{-6pt}
\end{center}%
}]

\begin{abstract}
\looseness=-1
We introduce \modelname{}, a \emph{compact spectral space} of body motion in video, in which each motion is a single code that can be compared and edited without training, transferred to a new subject, interpolated, composed, scaled in intensity and detail, or sampled anew. Our encoder expresses how a clip's DINOv2 features change over time in a fixed temporal basis, the discrete cosine transform (DCT), and compresses this change into 384 numbers. Because each frequency band is the same cosine pattern in every code, codes are aligned in time, ordered from coarse to fine, and can be blended or added without matching frames. What stays constant over a clip cancels, and keeping fewer frequencies leaves less of the performer and scene. The encoder is trained on real videos and on synthetic motions of several characters, and lightweight adapters let a frozen image-to-video model render any code with any subject. On unseen videos, our code ranks the same motion on other bodies higher than DisMo, a per-frame representation, while its nearest neighbors show the same performer about 40\% less often with a seven times more compact code. In generation, without per-motion training, \modelname{} transfers motion more faithfully than FlexiAct and a text-prompt baseline, at least as faithfully as DisMo on automatic metrics and clearly better between humanoid bodies, and its blends keep much more of both source motions than DisMo's. 
%
Users rank \modelname{} first nearly twice as often as DisMo for motion transfer (59\% vs.~30\%), place it top-two 94\% of the time, and prefer it in 76\% of the interpolation comparisons.
\vspace*{-12pt}%
\end{abstract}

\section{Introduction}
\label{sec:intro}

Recent advances in generative modeling have made image and video generation widely accessible. What a video adds to an image
is largely motion, and in many applications such as motion transfer, editing, and imitation
learning, what matters is the \emph{movement}, not who performs it. For example, a human squat should carry over to a
cartoon cat, as shown in Fig.~\ref{fig:teaser}.
This calls for a representation of motion that is disentangled from the performing body, and ultimately, a \emph{space},
in which
nearby points encode similar motions, and
simple operations on points, such as interpolation, produce natural changes in motion.
While images and language have such latent spaces \cite{radford2021learning,rombach2022high},
body motion in video does not.

Most existing motion specifications tend toward one of two extremes. Text is too coarse: for example, ``a person doing a cartwheel''
fixes neither timing nor amplitude. Reference-based methods are too literal, taking the motion from a source video as frame-aligned
signals such as pose, depth, or attention features \cite{zhang2023adding,ling2025motionclone}, or as weights and embeddings
optimized on one video \cite{kansy2025reenact,zhao2024motiondirector,zhang2025flexiact}.
These are records of a clip rather than descriptions of a motion, inheriting its length, timing, and often the performer.

Closest to a motion space is DisMo \cite{ressler2025dismo}, which extracts an appearance-invariant embedding of
general video motion in a single forward pass.
Its representation, however, remains a \emph{per-frame} sequence bound to the timing of the source video.
Since sequences of different clips share no temporal coordinates, they cannot be blended without matching frames,
and a variable-length sequence is not directly suited to sampling whole motions. On targets whose bodies differ strongly
from those in the source, DisMo tends to carry over the source's body (Fig.~\ref{fig:qual_adapt}).

\begin{figure*}[t]
  \centering
  \setlength{\abovecaptionskip}{3pt}%
  \includegraphics[width=\textwidth]{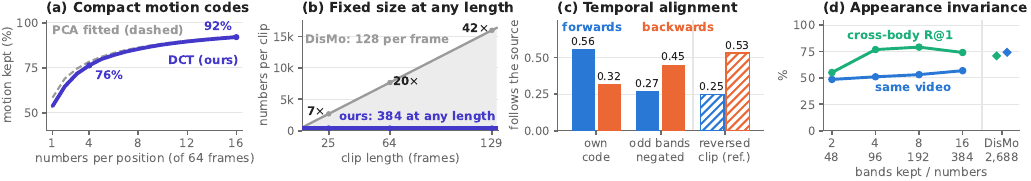}
  \caption{\textbf{Properties of the spectral code.} (a)~Share of DINOv2 feature change kept by the first $k$ DCT bands or $k$ fitted principal components
  (150 clips). (b)~Numbers per clip against clip length. (c)~Motion fidelity on the subject against the
  source played forwards and backwards (10 motions). (d)~Our code cut to its first $k$ bands, probed as in Table~\ref{tab:space} (simple motions: R@1\,$\uparrow$, same video\,$\downarrow$).}
  \label{fig:why}
  \vspace{-10pt}%
\end{figure*}

In this paper, we introduce \modelname{}, a \emph{compact spectral space} of body motion in video, in which each motion is
a single code representing its \emph{temporal} spectrum \emph{across video frames.}
Our encoder reads how the DINOv2 features \cite{oquab2024dinov2} of each image patch change over a clip and expresses
that change in a fixed temporal basis, the discrete cosine transform (DCT).
This gives the space three key properties:
\begin{itemize}
\item \emph{Compact motion codes.}
As bodies often move smoothly, the motion change can be well captured by a small number of low-frequency components (Fig.~\ref{fig:why}a).
Our code has a fixed size of 384 numbers at any clip length, whereas DisMo's per-frame sequence grows with it
(Fig.~\ref{fig:why}b).
\item \emph{Temporal alignment.}
With the spectral basis fixed, each frequency band is the same cosine pattern in every code, hence codes can be blended,
added, or reversed without matching video frames. For example, negating the odd
bands approximately plays a motion backwards (Fig.~\ref{fig:why}c).
\item \emph{Appearance invariance.}
As our encoder is trained on the temporal frequencies of feature change, what stays constant over a clip cancels, and keeping fewer frequencies
leaves fewer traces of the performer and scene (Fig.~\ref{fig:why}d). While DisMo draws its invariance from a limited
embedding size and strong augmentation, ours is rooted in the spectral structure and can be tuned post-training.
\end{itemize}

Our encoder is trained on synthetic motions rendered on several characters. A contrastive term pulls together the codes of one motion
across characters, and a decoder, used only in training, rebuilds how one character's features change from the code of another.
Real videos, which have no such pairs, contribute only to a reconstruction loss.

\begin{figure}[t]
  \centering
  \leavevmode\input{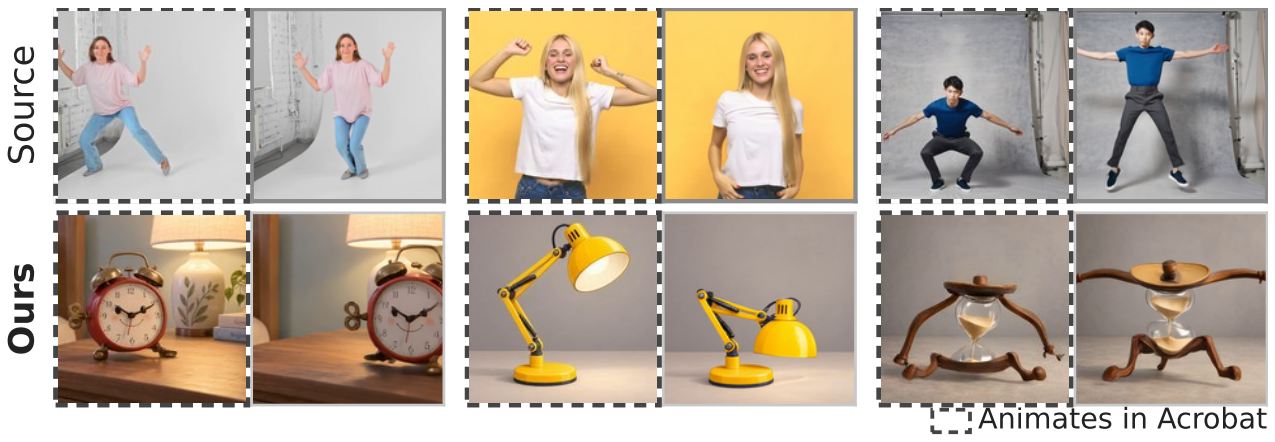}%
  \vspace{-18pt}%
  \caption{\textbf{Transfer onto very different targets.} Source and ours at the same moments. Side steps become an
  alarm clock side-stepping on a table, lowered arms make a desk lamp lower its head, and a crouch-and-extend transfers seamlessly onto an hourglass.}%
  \label{fig:qual_ours}
  \vspace{-12pt}%
\end{figure}

At inference, a frozen image-to-video model~\cite{wan2025wan} receives a \modelname{} code as its only motion signal, through lightweight adapters that expand it over the output frames with the inverse DCT.
As the generator receives only the code from the source video, nothing tells it what the source's body looked like. It takes the body from the target image instead and carries out the motion within that body's shape and proportions, be it a person, an animal, a toy, or a desk lamp (Figs.~\ref{fig:teaser},~\ref{fig:qual_adapt}, and~\ref{fig:qual_ours}). Any code, whether extracted, edited, or sampled, is rendered the same way.

\medskip
\noindent Our main contributions can be summarized as follows:
\begin{itemize}
\item \modelname{}, a compact spectral space of body motion in video, in which each motion is a fixed-size spectral code, largely invariant to the performer's appearance and aligned in time by a shared frequency basis.
\item A spectral encoder trained across characters, and lightweight adapters that let a frozen image-to-video model render any code with any subject.
\item Training-free operations in the space, including motion transfer, interpolation, composition, control of intensity and detail, and sampling of new motions.
\end{itemize}

We evaluate \modelname{} both directly and through generation.
On unseen videos, our code ranks the same motion performed by other bodies higher than DisMo while keeping less of the performer's appearance and being seven times more compact (Table~\ref{tab:space}).
In generation, without per-motion training, \modelname{} transfers motion more faithfully than the per-motion-tuned FlexiAct \cite{zhang2025flexiact}
and a text-prompt baseline; on automatic metrics it does at least as well as DisMo, including on DisMo's base model, and clearly better between humanoid bodies. Its blends also retain much more of both source motions than DisMo (Table~\ref{tab:transfer}). In a user study comparing our method to baselines, users rank \modelname{} first nearly twice as often as DisMo for transfer (59\% vs. 30\%), place it in their top two in 94\% of answers, and prefer it in 76\% of the interpolation comparisons.

Furthermore, our ablations show that, at the same code size, the DCT keeps the target intact better than learned or pooled-frame compression, follows body motion most closely when the target is itself a body, and is far more robust to camera shake (Table~\ref{tab:compression}, Fig.~\ref{fig:abl_hourglass}).

\section{Related Work}
\label{sec:related}

\paragraph{Motion representations.}
Videos have been factored into content and motion \cite{tulyakov2018mocogan}
and encoded by self-supervised models \cite{wang2023videomae}. Similar to
\modelname{}, DisMo \cite{ressler2025dismo} also builds on DINOv2 and
conditional LoRA, but conditions the video model on a sequence of per-frame
motion embeddings from the source, trained at scale on general motion. What Moves \cite{fundel2026whatmoves}
computes them for a chosen region, and X-NeMo \cite{zhao2025xnemo}
distills an identity-agnostic latent per frame for faces. 
All three keep one embedding per frame, tying a motion to its clip length and timing. In contrast, we represent motion 
by a single fixed-size code that can be blended and sampled.

\vspace{-8pt}

\paragraph{Motion transfer.}
Reference-based methods tune on a clip~\cite{zhao2024motiondirector} or follow its space-time 
features~\cite{ling2025motionclone,yatim2024space}.
For targets unlike the source, FlexiAct \cite{zhang2025flexiact} tunes an
action embedding on each reference video (1{,}500--3{,}000 steps per motion) and its ``frequency-awareness''
refers to denoising steps, not to motion over time. Reenact Anything \cite{kansy2025reenact} learns a motion-text embedding per motion,
Motion4Motion \cite{chen2026motion4motion} matches points between source and target and injects the source's dense motion flow into a
video transformer's self-attention, so the target follows the source's paths in the frame, and Motion Beyond Morphology
\cite{fang2026motion} trains a generator conditioned on the reference video
with cross-category pairs. These methods tune on or condition on each reference, 
leaving no code to blend or edit. In contrast, we encode a motion in one forward pass into a compact code that can
be blended, scaled and coarsened.
Frozen generators are also steered with learned inputs
\cite{gal2023image,lester2021power}, low-rank adapters \cite{hu2022lora} or
dense source-aligned signals such as pose or depth \cite{zhang2023adding}. Pose-driven animation \cite{hu2024animateanyone,tan2025animatex} transfers body motion through a skeleton, which targets without one, or with very different proportions, cannot follow.
Our conditional LoRA follows FiLM \cite{perez2018film} and reads a single compact code.

\vspace{-8pt}

\paragraph{Latent and spectral spaces of motion.}
Skeleton-based methods learn latent spaces of body motion from motion capture
and sample them with variational or diffusion models
\cite{petrovich2021action,ling2020character,tevet2023human}. They need a
skeleton and do not drive pixels. Learned spaces
are sampled through a variational objective or a separate model of their codes
\cite{ghosh2020variational,rombach2022high}, and novelty has been sought by
deviating from known styles \cite{elgammal2017can} or searching for it
directly \cite{lehman2011abandoning}. \modelname{} is learned from video, drives a
video model for bodies with no shared skeleton, and is searched for codes far
from known motions.
Frequency-domain descriptions of motion have a long history in animation
\cite{bruderlin1995motion,unuma1995fourier}. For skeletons, a DCT over time
encodes joint trajectories for motion prediction \cite{mao2019learning}, and
periodic autoencoders learn phase manifolds of body movement
\cite{starke2022deepphase}. Generative Image Dynamics \cite{li2024generative}
predicts a scene's oscillations from one image as per-pixel Fourier
coefficients. These describe skeleton joints or a scene's pixels; we use a
temporal DCT to encode the motion of any body in one code space.

\section{Method}
\label{sec:method}
\begin{figure*}[t]
  \centering
  \setlength{\abovecaptionskip}{3pt}%
  \includegraphics[width=\textwidth]{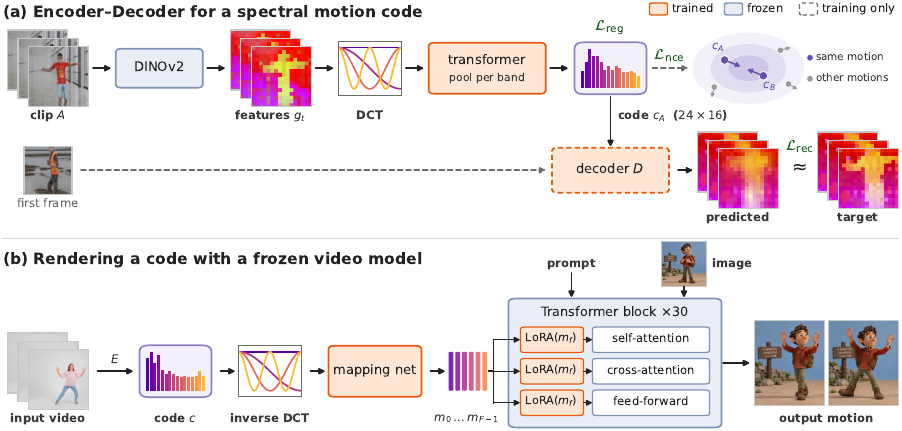}
  \caption{\textbf{\modelname Pipeline.} (a) The encoder turns the DINOv2 features of
  clip $A$ into one code $c_A$ through a temporal DCT, trained with a decoder that rebuilds the motion ($\mathcal{L}_{\text{rec}}$), a contrastive loss ($\mathcal{L}_{\text{nce}}$), and a regularizer that keeps the code's dimensions varied and decorrelated ($\mathcal{L}_{\text{reg}}$). (b) The inverse DCT and a mapping network expand the
  code into per-frame embeddings $m_0 \dots m_{F-1}$ that drive conditional LoRA layers in all 30 blocks of the frozen Wan2.2 transformer.}
  \label{fig:pipeline}
  \vspace{-10pt}%
\end{figure*}

Our aim is a compact code for the body motion in a video clip, largely independent of the performer's appearance and of the
clip's length, so that it can drive a frozen video diffusion model to render that motion with any subject.
To this end, a spectral encoder, trained with a decoder, compresses a clip's motion into a code $c \in \mathbb{R}^{M \times K}$ with $M = 24$
channels and $K = 16$ temporal frequencies (384 numbers; Fig.~\ref{fig:pipeline}a, Sec.~\ref{sec:encoder}). Lightweight adapters then
let the frozen model render the code with the subject of a still image (Fig.~\ref{fig:pipeline}b, Sec.~\ref{sec:conditioning}). The encoder is trained first, and the
adapters are then trained on its frozen codes. Operations on codes act on motions without further training
(Sec.~\ref{sec:operations}).

\subsection{Spectral Motion Encoder}
\label{sec:encoder}

\paragraph{Encoder.} 
All clips are resampled to $T=64$ frames, and each frame passes through a frozen DINOv2-B/14~\cite{oquab2024dinov2}, giving a $16\times16$ grid of patch features $g_t \in \mathbb{R}^{256\times768}$, $t=0,\dots,T-1$. A learned projection $P$ maps each feature to the transformer width of 384, and we subtract the first frame, $r_t = P(g_t)-P(g_0)$, so that what stays constant over the clip cancels. A temporal DCT at every patch keeps the $K=16$ lowest frequencies, $f_k = \frac{1}{T}\sum_t r_t B_{k,t}$ for $k<K$, where $B_{k,t}$ is the DCT basis over $T$ frames. A transformer with full attention processes the resulting $256\times K$ patch--frequency tokens. For each frequency $k$, its own learned query pools over the 256 patches, and a linear map projects the result to $M=24$ channels, giving column $c_{:,k}$. The code $c\in\mathbb{R}^{M\times K}$ acts as a pseudo-DCT of the motion. It is not the transform of any one signal, yet each column belongs to one frequency, ordered from slow to fast, and its inverse DCT gives $M$ trajectories over time.

\paragraph{Decoder.}
A decoder, used only in training, takes
the code and a first frame and predicts the DCT coefficients $\hat D_k$ of each
patch's displacement from the first frame, and the inverse DCT gives the features
$\hat g_t$ at every frame (Fig.~\ref{fig:pipeline}a).

\paragraph{Training and losses.}
Both are trained jointly with
\[
  \mathcal{L} = \mathcal{L}_{\text{rec}} + \lambda_{\text{nce}}\,\mathcal{L}_{\text{nce}}
              + \lambda_{\text{reg}}\,\mathcal{L}_{\text{reg}}.
\]
The reconstruction loss $\mathcal{L}_{\text{rec}}$ compares predicted and true displacements from the first frame
through their DCT coefficients. Each frequency's error is divided by its own energy, which gives the fast
frequencies a boost, since the slowest one holds about half of the energy. We also add two auxiliary terms, found
empirically to refine the reconstruction: a squared error weighted toward patches that move, and a cosine distance
between the two displacements.

The reconstruction task depends on the data. In synthetic data the same motion is performed by several characters, so
reconstruction runs across characters. For clips $A$ and $B$ of the same motion, the
decoder receives $A$'s code and $B$'s first-frame features and must predict
$B$'s features at every frame, and symmetrically from $B$ to $A$. Real clips have no such pairs, so each is
seen in two augmented views, and the decoder predicts each view from the other view's code. $\mathcal{L}_{\text{nce}}$ is an InfoNCE loss over the synthetic codes of a batch, in which
the positive for each code is the other character's code of the same motion. It pulls
the codes of one motion together across bodies and pushes other motions away.
$\mathcal{L}_{\text{reg}}$ keeps the code's dimensions varied and decorrelated.

\subsection{Rendering with a Frozen Video Model}
\label{sec:conditioning}

We pair the encoder with a fully frozen Wan2.2-TI2V-5B \cite{wan2025wan} and trainable adapters, a mapping network and conditional LoRA layers. The training prompt is the same for
every clip, ``an articulated motion, high quality''. At inference, a description of the target's appearance is placed before it.

\paragraph{Inverse DCT and per-frame expansion.}
The code, normalized to a fixed norm, is expanded over the $F$ latent
frames of the output ($F = 16$ in training) with the inverse DCT (the same basis with $T = F$, so frequency $k$
means the same thing as in the encoder),
\[
  u_f = \sum_{k} c_{:,k} B_{k,f} \in \mathbb{R}^{M}, \qquad f = 0,\dots,F-1,
\]
where $B_{k,f}$ is the value of the $k$-th DCT cosine at latent frame $f$.
At other lengths, the cosines are sampled at $F$ evenly spaced latent frames, so the output length, not the source, sets the motion's absolute speed.
A mapping network turns each $u_f$ into an embedding
$m_f \in \mathbb{R}^{768}$. We center its output with a running mean and a
running scale, fixed at test time. Without centering we found the network drifts
toward emitting the same embedding for every clip.
These per-frame embeddings come from the code alone, not from the source.

\paragraph{Conditional LoRA.}
We add a conditional LoRA \cite{hu2022lora,ressler2025dismo} to linear layers
throughout the transformer. For a frozen layer $y = Wx$ and a token $x$ in
latent frame $f$,
\[
  y = Wx + U\big[(1 + \gamma(m_f)) \odot Vx + \beta(m_f)\big],
\]
where $V$ and $U$ are the low-rank factors, and $\gamma$ and $\beta$ are
linear maps of the frame's motion embedding. $U$, $\gamma$ and $\beta$ start at zero, so training starts from the
unmodified model.

\paragraph{Training and sampling.}
We train the mapping network and LoRA with the flow-matching loss, replacing the code with a learned null code
for $10\%$ of samples, for classifier-free guidance \cite{ho2022classifier} between the code and the null code.
\subsection{Operations in the Motion Space}
\label{sec:operations}

Every operation below builds a code from unit-norm codes $\hat c = c / \lVert c \rVert$; nothing needs training or fine-tuning (Fig.~\ref{fig:qual_ops}).

As codes are vectors, motions can be combined through simple arithmetic.
The blend $(1-w)\,\hat c_A + w\,\hat c_B$, renormalized, \emph{interpolates} from motion
$A$ to motion $B$ (Fig.~\ref{fig:qual_ops}, Interp.\ row), and the sum $\hat c_A + \hat c_B$, the blend at $w = 0.5$ up to scale, \emph{composes} them (Fig.~\ref{fig:qual_ops}, $w = 0.5$).
The direction $d$ from the mean code of the $20\%$ stillest training clips to that of the $20\%$ most
moving ones controls \emph{intensity}: $\hat c + \alpha d$ turns a motion down ($\alpha < 0$) or up
($\alpha > 0$) (Fig.~\ref{fig:qual_ops}, Intensity row).

Keeping only the first $K' < K$ frequency bands gives a \emph{coarser} version of a
motion, its slow course without the fast detail, which sets how closely a target follows the source (Fig.~\ref{fig:qual_ops}, Coarse row).
Since even DCT bands are symmetric in time and odd ones are antisymmetric, negating
the odd columns approximately plays a motion backwards (Fig.~\ref{fig:why}c).

Finally, the space holds motions that no clip performed. Codes sampled from a Gaussian $\mathcal{N}(\mu, \Sigma)$ fitted to the unit
training codes are new motions. To push further from the data, we use novelty search \cite{lehman2011abandoning}.
Starting from real codes, we ascend novelty $\nu$, one minus the mean cosine similarity to the three nearest training codes,
penalized for leaving the support of the data, and keep a diverse subset of the most novel codes inside it.

\subsection{Implementation Details}
\label{sec:impl}

We build on Wan2.2-TI2V-5B \cite{wan2025wan} as it generates different video lengths, keeps the target's aspect ratio and has 16 latent frames, one per band of our code. For the baseline comparisons, we also build on CogVideoX-5B-I2V \cite{yang2024cogvideox}, as both DisMo and FlexiAct build on it.

\section{Experiments}
\label{sec:experiments}

We test how well the motion space separates motion from appearance (Sec.~\ref{sec:exp_space}), whether a frozen generator reproduces a motion from the code alone, and whether operations on codes behave like operations on motions (Sec.~\ref{sec:exp_transfer}).

\begin{figure*}[t]
  \centering
  \leavevmode\input{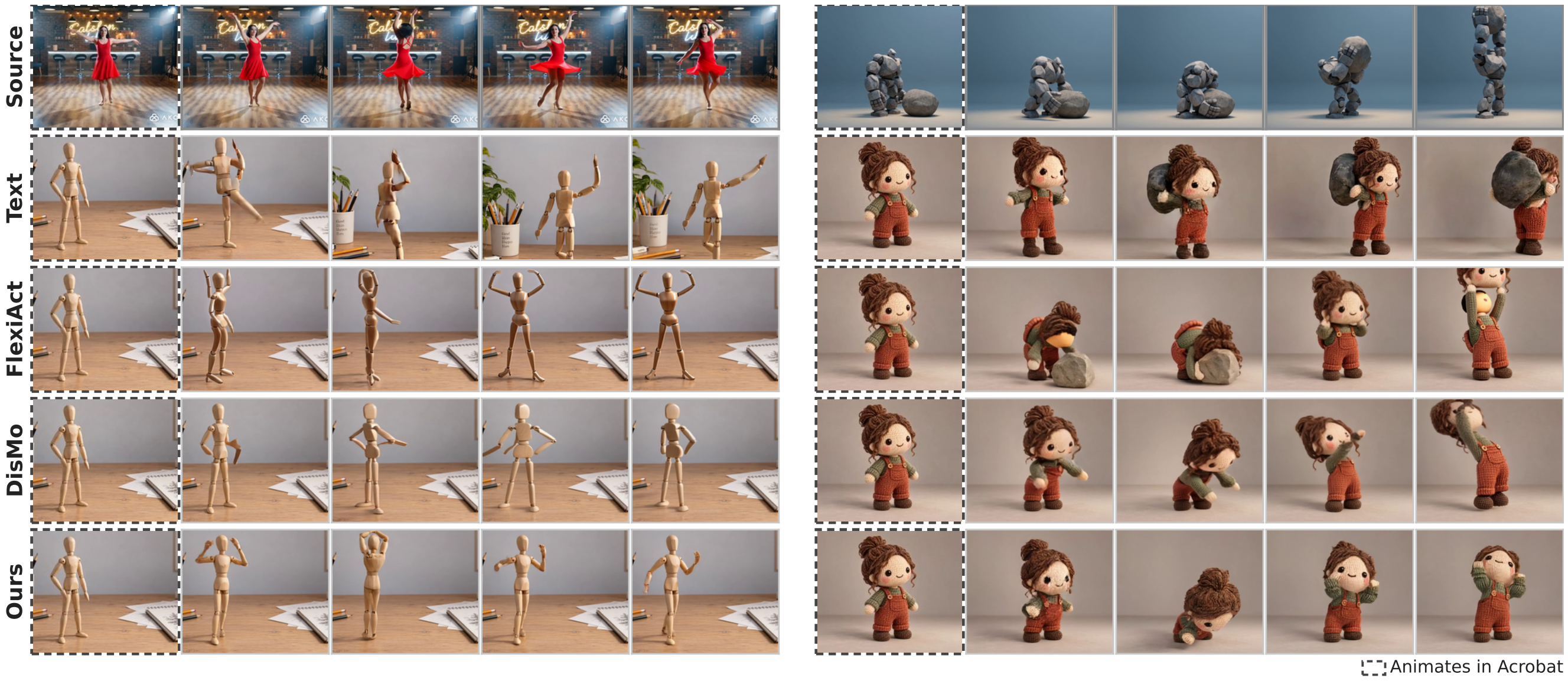}
  \vspace{-18pt}%
  \caption{\textbf{The motion is adapted to the target.} A salsa twirl transferred to a wooden mannequin (left) and a
  stone golem lifting a boulder transferred to a knitted doll (right), by the text baseline, FlexiAct \cite{zhang2025flexiact},
  DisMo \cite{ressler2025dismo} and ours. Ours performs each motion with the
  target's own body, whereas the baselines carry over the source's shape or add its boulder. Animates in Adobe Acrobat.}%
  \label{fig:qual_adapt}
  \vspace{-9pt}%
\end{figure*}
\subsection{Setup}
\label{sec:setup}

\paragraph{Data.}
Our synthetic dataset has 1{,}704 animations, each rendered on 5 of 107 characters
(8{,}520 clips). Real video comes from IDEA400, part of Motion-X \cite{lin2023motionx} (12{,}037 clips), and
HAA500 \cite{chung2021haa500} (6{,}944 clips), resampled or cut to 64 frames.
The encoder reads them at $224 \times 224$, as a full frame and a tight crop around
the motion. Real clips train it only through
reconstruction (Sec.~\ref{sec:encoder}). The adapters train on all 3 sets.

\vspace{-5pt}
\paragraph{Evaluation data.}
Test videos and images are unseen in training. The motion space is probed
on 253 segments of 59 real videos (the out-of-distribution, or OOD set) and on 2 \emph{motion grids}
generated with Wan2.2-T2V-A14B \cite{wan2025wan}, one prompt per clip. 6 characters
(people, a robot, a doll, an ogre) perform 12 simple motions in one and 12
two-step motions in the other (72 clips each). An LLM writes the prompts, asked only for different motions on different subjects. Labels come from the prompts, so timing
and poses vary, and some two-step clips miss the prompted order.
As FlexiAct tunes for ${\sim}4$\,h per motion (H200), the baseline transfer comparison favors few motions on diverse targets. 12 source videos of different kinds (6 OOD and 6 two-step motion-grid clips) are rendered on 10 target
images from our 101 still images of people, animals, figures and objects. Interpolation is scored on 150 blends of two body motions from unseen videos, each on a target image. Its user study uses 10 motion pairs generated with Wan2.2-T2V-A14B, drawn at random.

\vspace{-5pt}
\paragraph{Baselines.}
For the motion space, we compare with DisMo's released motion extractor \cite{ressler2025dismo}, whose code is its full sequence of per-frame embeddings. For transfer, we compare with DisMo, FlexiAct \cite{zhang2025flexiact} and a text baseline; for interpolation, with DisMo. DisMo and FlexiAct use CogVideoX-5B \cite{yang2024cogvideox}, so we report \modelname{} on both Wan2.2 and CogVideoX-5B (Sec.~\ref{sec:impl}); the user study uses CogVideoX-5B. The text baseline is Wan2.2-TI2V-5B \cite{wan2025wan} without adapters. Its prompt describes the source motion in order: which body parts move, in which direction, how often and how fast. Qwen2.5-VL-32B \cite{bai2025qwen25vl} writes all prompts, and appearance prompts describe only the target image.

\subsection{Motion Space Analysis}
\label{sec:exp_space}

Following DisMo \cite{ressler2025dismo}, Table~\ref{tab:space} probes both motion spaces without generation, by nearest-neighbor retrieval on videos neither encoder saw in training. On the 2 motion grids (Sec.~\ref{sec:setup}),
\emph{cross-body} retrieval asks whether a clip's nearest neighbor among the other characters performs the same motion (R@1). mAP measures
how highly the 5 other characters' versions of the same motion rank. \emph{Same character} takes a clip's 5 nearest neighbors in its
grid and, among those performing another motion, counts the share showing the same character. Every character performs every
motion, so only appearance makes them close. On the OOD set, \emph{same video} is how often a segment's nearest neighbor comes from
its own source video, which shares the performer, clothing, scene and camera but not always the motion. Lower is better for both. Our code finds the same motion on another body in 3 of 4 cases and ranks its other versions higher than DisMo, with a
seventh of the numbers. Its nearest neighbors show the same character about $40\%$ less often ($22$--$25\%$ vs.\ $37$--$42\%$, chance is $17\%$)
and come from the same video less often ($57\%$ vs.\ $74\%$).
\begin{table}[t]
\centering
\footnotesize
\setlength{\tabcolsep}{2.2pt}
\setlength{\aboverulesep}{1pt}\setlength{\belowrulesep}{1pt}\renewcommand{\arraystretch}{1.0}\setlength{\abovecaptionskip}{4pt}%
\begin{tabular*}{\columnwidth}{@{\extracolsep{\fill}}lcccccccc@{}}
\toprule
& & \multicolumn{2}{c}{R@1\,$\uparrow$} & \multicolumn{2}{c}{mAP\,$\uparrow$} & \multicolumn{2}{c}{same char.\,$\downarrow$} & same \\
\cmidrule(lr){3-4}\cmidrule(lr){5-6}\cmidrule(lr){7-8}
& size\,$\downarrow$ & simple & 2-step & simple & 2-step & simple & 2-step & video\,$\downarrow$ \\
\midrule
DisMo \cite{ressler2025dismo} & 2{,}688 & $71\%$ & $\mathbf{73\%}$ & $.583$ & $.531$ & $37\%$ & $42\%$ & $74\%$ \\
\modelname{}          & $\mathbf{384}$ & $\mathbf{74\%}$ & $\mathbf{73\%}$ & $\mathbf{.602}$ & $\mathbf{.613}$ & $\mathbf{22\%}$ & $\mathbf{25\%}$ & $\mathbf{57\%}$ \\
\bottomrule
\end{tabular*}
\caption{\textbf{Motion space}, probed without generation on unseen data
(Sec.~\ref{sec:exp_space}). R@1, mAP and same char.\ on each motion grid (simple and two-step motions), each clip shrunk to a random 50--100\% and
placed at random, mean of 3 framings. Same video: OOD set.}
\label{tab:space}
  \vspace{-8pt}%
\end{table}

\begin{table*}[t]
\centering
\setlength{\abovecaptionskip}{4pt}
\setlength{\tabcolsep}{1.5pt}
\providecommand{\dottedrow}[1]{\noalign{\vskip .4ex\hbox to #1{\leaders\hbox to 2pt{\hss\vrule height .4pt depth 0pt width .5pt\hss}\hfill}\vskip .65ex}}
\begingroup\footnotesize
\setlength{\aboverulesep}{1pt}\setlength{\belowrulesep}{1pt}\renewcommand{\arraystretch}{1.0}%
\begin{tabular*}{\textwidth}{@{\extracolsep{\fill}}lcccccccccccccccc}
\toprule
 & & & \multicolumn{4}{c}{transfer, 120 pairs} & \multicolumn{3}{c}{transfer, 300 pairs} & \multicolumn{2}{c}{body$\to$body, 107} & \multicolumn{3}{c}{interpolation, 150} & \multicolumn{2}{c}{body$\to$body, 71} \\[-.3ex]
\cmidrule(lr){4-7}\cmidrule(lr){8-10}\cmidrule(lr){11-12}\cmidrule(lr){13-15}\cmidrule(lr){16-17}
method & frames & time\,$\downarrow$ & mot.\,$\uparrow$ & temp.\,$\uparrow$ & img.\,$\uparrow$ & user 1st/top-2\,$\uparrow$ & mot.\,$\uparrow$ & temp.\,$\uparrow$ & img.\,$\uparrow$ & mot.\,$\uparrow$ & img.\,$\uparrow$ & mot.\,$\uparrow$ & temp.\,$\uparrow$ & user\,$\uparrow$ & mot.\,$\uparrow$ & temp.\,$\uparrow$ \\
\midrule
text baseline                       & 61 & $\mathbf{29}$\,s & $.256$ & $.971$ & $.854$ & $4/19$ & $.307$ & $.965$ & $.841$ & $.410$ & $.850$ & -- & -- & -- & -- & -- \\
FlexiAct \cite{zhang2025flexiact}   & 49 & $3.9$\,h & $.361$ & $.973$ & $.858$ & $7/19$ & -- & -- & -- & -- & -- & -- & -- & -- & -- & -- \\
DisMo \cite{ressler2025dismo}       & 25 & $53$\,s & $.378$ & $.982$ & $.908$ & $30/67$ & $.371$ & $.983$ & $.910$ & $.360$ & $.896$ & $.226$ & $\mathbf{.990}$ & $24\%$ & $.237$ & $\mathbf{.988}$ \\
\modelname{} (Cog-5B) & 49 & $119$\,s & $\mathbf{.421}$ & $\mathbf{.984}$ & $\mathbf{.917}$ & $\mathbf{59/94}$ & $.380$ & $\mathbf{.985}$ & $\mathbf{.918}$ & $.418$ & $\mathbf{.899}$ & $.322$ & $.987$ & $\mathbf{76\%}$ & $.370$ & $.983$ \\
\dottedrow{\textwidth}
\modelname{} (Wan-5B)       & 61 & $40$\,s & $.410$ & $.983$ & $.905$ & -- & $\mathbf{.394}$ & $.983$ & $.902$ & $\mathbf{.421}$ & $.888$ & $\mathbf{.348}$ & $.986$ & -- & $\mathbf{.384}$ & $.981$ \\
\bottomrule
\end{tabular*}
\endgroup
\caption{\textbf{Motion transfer and interpolation} (Sec.~\ref{sec:exp_transfer}). Mot.: motion fidelity; img.: CLIP
similarity to the target image. Time: per video on one H200, including any per-motion training but not the text baseline's captioning. User 1st/top-2: \% of
answers (user study, 10 examples) that rank the method first / first or second of the four. User (interpolation): share that choose the method of the two (user study, 10 pairs). Interpolation: in-between videos
($w = .25, .5, .75$), motion scored against both source motions. FlexiAct not run on 300 pairs.}
\label{tab:transfer}
  \vspace{-4pt}
\end{table*}

\begin{table}[t]
\centering
\footnotesize
\setlength{\tabcolsep}{2pt}
\setlength{\aboverulesep}{1pt}\setlength{\belowrulesep}{1pt}\renewcommand{\arraystretch}{1.0}\setlength{\abovecaptionskip}{3pt}%
\begin{tabular*}{\columnwidth}{@{\extracolsep{\fill}}lll@{}}
\toprule
operation & measure & result \\
\midrule
coarse to fine & source found among 21, $k < 2/4/8/16$ & $23/55/88/91\%$ \\
intensity & motion in body heights, $\alpha = -2 \rightarrow +2$ & $.05 \rightarrow 1.38$ \\
composition & cosine to A\,/\,B: A+B vs.\ the other alone & $.34/.33$ vs.\ $.06$ \\
sampling & temp.: sampled vs.\ same-target transfers & $.978$ vs.\ $.969$ \\
\bottomrule
\end{tabular*}
\caption{\textbf{Operations, measured} on 168 pairs, 105 sums, 210 samples (chance $5\%$).}
\label{tab:ops_main}
\vspace{-14pt}%
\end{table}

\begin{figure*}[t]
  \centering
  \leavevmode\input{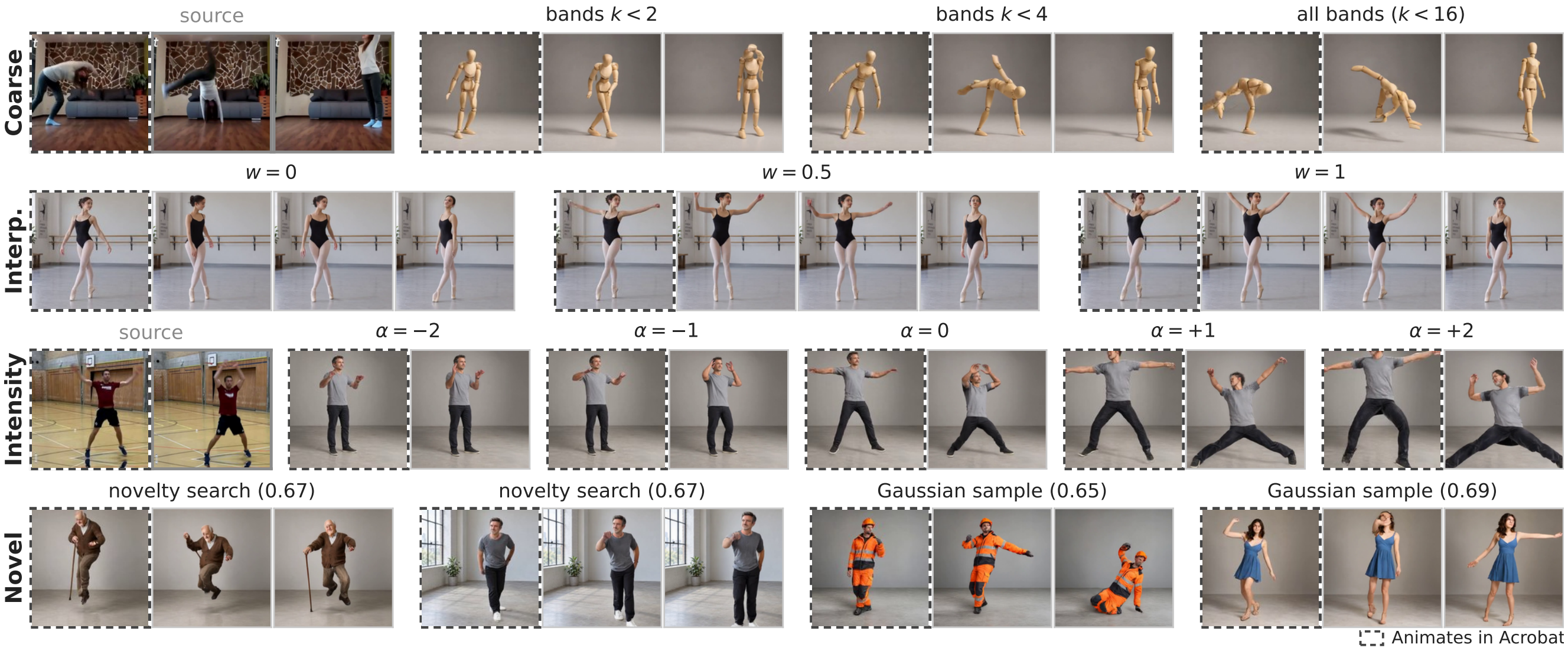}%
  \vspace{-18pt}%
  \caption{\textbf{Operations in the code space.} \emph{Coarse:} a
  cartwheel's code keeps only its bands $k < 2$, $4$ or all 16: 2 bands only lean and reach up, 4 add a deep bend with
  a kick, all 16 do the whole cartwheel. \emph{Interp.:} blending the codes of a sway and an arm raise with weight $w$: the dancer only sways at $w = 0$,
  combines both at $w = 0.5$ and only raises her arms at $w = 1$. \emph{Intensity:} a jumping jack moved along the intensity direction
  from $\alpha = -2$ to $+2$ goes from barely raising the arms to a wide star jump. \emph{Novel:} codes with no source
  video, from novelty search (first two) or the Gaussian over codes (last two); in brackets each code's novelty $\nu$
  (training median $0.32$).}
  \label{fig:qual_ops}
  \vspace{-8pt}
\end{figure*}
\subsection{Motion Transfer and Interpolation}
\label{sec:exp_transfer}

On the 120 transfer pairs, we use DisMo's metrics, motion fidelity and temporal consistency, which depend on neither motion encoder.
\emph{Motion fidelity} follows Yatim \etal \cite{yatim2024space}: points are
tracked with CoTracker \cite{karaev2024cotracker} in both videos, each generated
track is matched to the source track whose frame-to-frame directions correlate
best with it, and the correlations are averaged. As our scope is the body's motion, we compute it
on tracks with camera motion removed and weight it by how far the subject moves relative to the source, capped at 1.
\emph{Temporal consistency} is the CLIP \cite{radford2021learning} similarity of
consecutive frames, and \emph{image similarity} that of each frame to the target image. All metrics are computed on 25 evenly spaced frames of each video, the length of DisMo's output, which slightly favors DisMo. Confidence intervals (CI, 95\%) come from 10{,}000 bootstrap resamples of the pairs, or of the participants for the user study.
In Table~\ref{tab:transfer}, \modelname{} follows the source motion more closely than all baselines ($0.42$ on
CogVideoX and $0.41$ on Wan2.2, against $0.38$ for DisMo; $+0.04$ on CogVideoX, CI $[0.02, 0.07]$) at similar temporal consistency and image similarity, without per-motion training ($119$\,s per
video on CogVideoX and $40$\,s on Wan2.2, against $3.9$\,h for FlexiAct).
On 300 further body-motion samples from videos unseen in training (the OOD set, the two-step motion grid and other action clips),
DisMo scores $0.37$ against $0.38$ for \modelname{} on CogVideoX and $0.39$ on Wan2.2 (Table~\ref{tab:transfer}; $+0.01$ on CogVideoX, CI $[-0.01, 0.03]$). The gap is larger for the 107 that
transfer motion to a humanoid body (body$\to$body), which covers people but also figures with very distinct proportions, such as dolls, a paper boy or a scarecrow: $0.42$ and $0.42$ against $0.36$ ($+0.06$ on CogVideoX, CI $[0.03, 0.09]$). Although our cross-character pairs are synthetic, the gain carries over to the 100 OOD samples with real source videos ($+0.05$ on CogVideoX, CI $[0.01, 0.09]$). We attribute this to real videos making up $69\%$ of the training clips and to the small code, which leaves little capacity for appearance.

Automatic metrics miss how convincing a motion is, so we also run a user
study (Table~\ref{tab:transfer}; 29 participants for transfer and 27 for interpolation, 10 examples each). Users clearly prefer our method: they rank it first in $59\%$ of transfer answers (CI $[53\%, 66\%]$, chance $25\%$), $30$ points more often than DisMo (CI $[18, 41]$), and choose it in $76\%$ of interpolation answers (CI $[70\%, 82\%]$). On non-human targets, the preference is strongest ($76\%$ of first places), while on those of the 300 samples the motion score ties. In Fig.~\ref{fig:qual_adapt}, DisMo turns the golem's boulder into the doll's stretched head and keeps the mannequin facing forward instead of twirling. The text baseline and FlexiAct, whose prompts describe the source's action, add the boulder itself. Our generator never sees the source, so the mannequin twirls and the doll bends, keeping their own proportions.

For interpolation, we blend 2 codes (for DisMo, its embeddings frame by frame). This changes the
motion steadily from one code to the other (Figs.~\ref{fig:teaser},
\ref{fig:qual_ops}). Our blends keep the motion of both sources, on average $64\%$ of its amount on CogVideoX against $40\%$ for DisMo,
whose blends lose much of it (Table~\ref{tab:transfer}; $+0.10$ on CogVideoX, CI $[0.07, 0.12]$), most clearly on the 71 blends onto a humanoid body
(body$\to$body: $0.37$ and $0.38$ against $0.24$; $+0.13$ on CogVideoX, CI $[0.10, 0.16]$).
The other operations of Sec.~\ref{sec:operations} behave as expected (Table~\ref{tab:ops_main}). We re-encode each generated video with our encoder and compare it with the 21 source codes: sums lie close to both motions, and codes cut to fewer bands still identify their source. On tracks, the subject's motion grows with $\alpha$ in every pair, and sampled codes are as temporally consistent as real transfers.

\section{Ablations}
\label{sec:ablations}

\begin{figure}[t]
  \centering
  \leavevmode\input{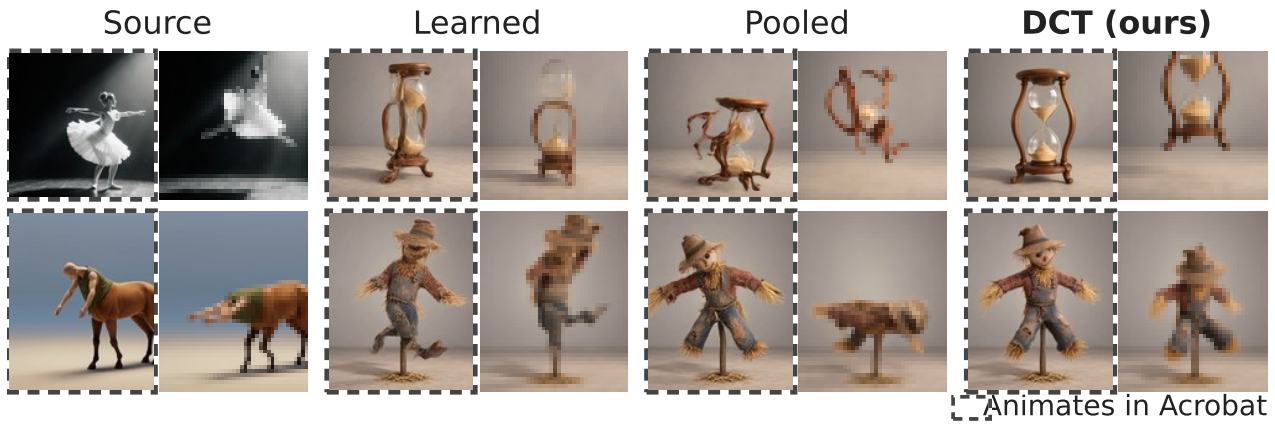}%
  \vspace{-16pt}%
  \caption{\textbf{Temporal compression}, same moments.
  Top: a ballet jump on an hourglass; bottom: a centaur's bow on a scarecrow.
  Learned compression distorts the subject, pooled frames break it apart or fold it flat, and the DCT keeps it whole while it moves.}%
  \label{fig:abl_hourglass}
  \vspace{-4pt}   
\end{figure}

We ablate compression techniques, the temporal basis, and the encoder's design. Due to compute budget, ablations train from scratch for fewer steps than the final model.
\vspace{-8pt}
\paragraph{Temporal compression.}
Transfer is scored on body motion in 320 pairs from 51 unseen clips (30 real, 21 generated), 198 with a person or humanoid source (body$\to$all), 55 of them with a
humanoid target (body$\to$body). Learned queries compressing the clip's motion (learned compression) collapse the code. Its slots become near-copies, and 21
principal components hold $90\%$ of its variance (Table~\ref{tab:compression}). A low-pass filter averaging every 4 frames before the
transformer (pooled frames) follows body motion about as well as the DCT ($0.387$ vs.\ $0.382$), less closely when the target is a humanoid ($0.383$ vs.\ $0.415$, DCT $+0.033$, CI $[0.001, 0.060]$), and keeps the target
less well ($0.873$ vs.\ $0.885$; DCT $+0.012$, CI $[0.006, 0.019]$). It breaks the subject apart or folds it flat, while the DCT keeps it whole and moving
(Fig.~\ref{fig:abl_hourglass}). The DCT keeps a motion's \textit{essence} rather than copying the source frame by frame, with coarse-to-fine control for free (Fig.~\ref{fig:qual_ops}). It also copes better with camera shake, which pooled frames fold into
their slots. With the query shaken or vibrating by up to $3$--$4\%$ of the frame, cross-body R@1 on both grids, without Table~\ref{tab:space}'s random resizing, is $53\%$/$64\%$ for the DCT vs.\ $37\%$/$40\%$ for pooled frames, which lead only when clean ($68\%$ vs.\ $63\%$).\par%

\paragraph{Temporal basis.}
Without training, 16 of 64 DCT bands retain nearly as much motion as the best
fitted basis (Fig.~\ref{fig:why}a). Without a time axis (the same code at every frame), an
adapter barely moves the subject (motion fidelity $0.17$ vs.\ $0.33$).
Fewer bands retain less appearance and video-specific detail, yet cross-body
retrieval holds down to 4 bands before dropping sharply (Fig.~\ref{fig:why}d).

\begin{table}[t]
\vspace{1pt}%
\centering
\footnotesize
\setlength{\tabcolsep}{1.2pt}
\setlength{\aboverulesep}{1pt}\setlength{\belowrulesep}{1pt}\renewcommand{\arraystretch}{1.0}\setlength{\abovecaptionskip}{4pt}%
\begin{tabular*}{\columnwidth}{@{\extracolsep{\fill}}lccccccccc}
\toprule
 & \multicolumn{2}{c}{body$\to$all} & \multicolumn{2}{c}{body$\to$body} & \multicolumn{2}{c}{code} & \multicolumn{3}{c}{R@1 (\%)} \\
\cmidrule(lr){2-3}\cmidrule(lr){4-5}\cmidrule(lr){6-7}\cmidrule(lr){8-10}
 & mot.$\uparrow$ & img.$\uparrow$ & mot.$\uparrow$ & img.$\uparrow$ & dims$\uparrow$ & corr.$\downarrow$ & clean$\uparrow$ & shake$\uparrow$ & vib.$\uparrow$ \\
\midrule
learned             & $.291$ & $.848$ & $.324$ & $.842$ & $21$ & $.96$ & $48$ & $26$ & $24$ \\
pooled              & $\mathbf{.387}$ & $.873$ & $.383$ & $.864$ & $75$ & $.81$ & $\mathbf{68}$ & $37$ & $40$ \\
DCT (ours)          & $.382$ & $\mathbf{.885}$ & $\mathbf{.415}$ & $\mathbf{.868}$ & $\mathbf{112}$ & $\mathbf{.39}$ & $63$ & $\mathbf{53}$ & $\mathbf{64}$ \\
\bottomrule
\end{tabular*}
\caption{\textbf{Temporal compression}. Mot.: motion fidelity; img.: CLIP similarity to the target;
dims: components holding $90\%$ of the code's variance; corr.: mean $|$correlation$|$ of neighboring slots; R@1: cross-body retrieval with a clean, shaken or vibrating query.}
\label{tab:compression}
  \vspace{-14pt}%
\end{table}

\paragraph{Encoder design.}
Removing $\mathcal{L}_{\text{nce}}$ or replacing DINOv2 with RGB patches costs the most cross-body retrieval (R@1 $66\%$/$42\%$ against $75\%$), even on synthetic training pairs ($83\%$/$82\%$ against $100\%$). The per-frequency balance and the frame-0 subtraction mainly help on fast motions (mAP on the fastest third $0.455$ against $0.443$ and $0.435$) and keep less of what is specific to a video (same video $56\%$ against $62\%$ and $64\%$ without them).

\section{Conclusion}

We introduced \modelname{}, a compact spectral space of body motion in video. Each motion is a single code of 384 numbers, the temporal spectrum of how a clip's features change, in a fixed DCT basis. Its codes are aligned in time and ordered from coarse to fine, so motions can be blended, added, scaled and sampled without
matching frames. A frozen image-to-video model renders any code on any subject,
transferring motion at least as faithfully as per-frame baselines, clearly better between humanoid subjects, and more faithfully than per-motion ones. Blends keep the motion of both sources, which per-frame embeddings lose.

By construction, the code cannot register timing below one
latent frame, and generated videos
can soften sharp timing. Since every clip is resampled to 64 frames, a code describes a motion relative to its clip. The same squat done twice lands in higher bands than done once, absolute speed comes from the output length (Sec.~\ref{sec:conditioning}), and a blend does not align the phases of its two motions. Despite better robustness to camera shake (Sec.~\ref{sec:ablations}), strong camera motion in the source can still carry over into the result. We mainly test with body motion, where the gain over per-frame methods is largest (body$\to$body, Table~\ref{tab:transfer}). A learned
prior over the DCT codes is a natural next step, to keep blends on plausible motions
rather than straight lines between codes, and to find novel codes
beyond a Gaussian.

{
    \small
    \bibliographystyle{ieeenat_fullname}
    \bibliography{main}
}

\end{document}


\raggedbottom
\makeatletter
\setlength{\@fptop}{0pt}\setlength{\@fpsep}{14pt plus 2pt}\setlength{\@fpbot}{0pt plus 1fil}
\setlength{\@dblfptop}{0pt}\setlength{\@dblfpsep}{14pt plus 2pt}\setlength{\@dblfpbot}{0pt plus 1fil}
\makeatother

\clearpage
\setcounter{page}{1}   
\maketitlesupplementary
\appendix

This supplementary material details the networks, their training, the operations on codes, the evaluation protocols, the
baselines and the user studies (Sec.~\ref{sec:supp_impl}), and gives additional ablation details (Sec.~\ref{sec:supp_abl}). Sec.~\ref{sec:supp_res}
shows interpolation and composition, non-body motion and failure cases, and what nearest neighbors share in DisMo's space and
ours, measures the operations of Sec.~3.3 and gives further examples of each, and ends
with the limitations of the code.

\section{Implementation Details}
\label{sec:supp_impl}

\subsection{Networks and Training}

\paragraph{Encoder--decoder.}
Each clip is resized to $224 \times 224$, as the full frame or as one square window around its moving region, fixed over
the clip, with equal probability. The encoder and decoder transformers each have 4 layers with 6 heads and full attention
over the $16 \times 256$ tokens (8.1M and 7.8M parameters). Each decoder token is built from a code column $c_{:,k}$ and
$g_0$ (first frame) at that patch, and the trajectory is anchored at the first frame,
$\hat g_t = g_0 + \sum_{k} \hat D_k \,(B_{k,t} - B_{k,0})$. The two views of a real clip each take a random window of at
least $70\%$ of the frame, a rotation within $\pm 8^\circ$, color jitter of $\pm 30\%$ and, with probability 0.3, a
Gaussian blur. They never differ in timing and are never mirrored.
\textbf{Training:}
Each step uses 12 synthetic clip pairs and, once real video is added, 12 real
clips. We use AdamW without weight decay, a cosine schedule, gradient clipping
at 1.0 and bf16. $\lambda_{\text{nce}} = 0.3$ with temperature 0.1,
and $\lambda_{\text{reg}} = 1$, with $\mathcal{L}_{\text{reg}}$ weighting the variance floor by 0.1 and the
decorrelation by 1. Training runs for approximately 200k steps, starting with synthetic and gradually adding real videos.

\paragraph{Adapters.}
The mapping network is a linear map from 24 to 768 dimensions followed by
2 residual blocks (RMSNorm, GEGLU, $3\times$ expansion). The LoRA has rank 64 and is applied in all 30 blocks to the self-attention
projections, the feed-forward layers, and the cross-attention query and output (240 layers, 172M trainable parameters
with the mapping network). Since the embeddings are centered, $\gamma$, $\beta$ and $V$ have bias terms. We train with
AdamW at learning rate $10^{-4}$ ($10^{-3}$ for the null code), 40 to 162 clips per update, noise levels
$t \sim \mathcal{U}[0.3, 1]$ and an exponential moving average of 0.999 for about 500k clip samples at $256 \times 256$, with synthetic, IDEA400 and HAA500 clips drawn at $40/30/30\%$. We then fine-tune for 70k
samples in three shapes ($640 \times 640$, $832 \times 480$ and $704 \times 576$).

\paragraph{CogVideoX version.}
For the baseline comparison (Table~2) and the user study, we also adapt a frozen CogVideoX-5B-I2V with a rank-128 conditional LoRA (rank chosen in preliminary tests) on the attention projections and feed-forward layers of every block, conditioned per latent frame as in Sec.~3.2. Its position embeddings are learned for 49 frames, so we train and sample at 49 frames and $720 \times 480$ (13 latent frames), and the code keeps only its first 13 bands. The 3 dropped bands hold about $2\%$ of the motion energy on average, but up to $21\%$ on fast repeated motions such as jumping jacks; this is one reason we use Wan, with its 16 latent frames, as our main model. Prompt, null code, data mix and optimizer follow the Wan adapters. We train on 125k clips with $t \in [300, 1000]$.

\subsection{Motion Space}
\paragraph{Sampling and novelty search.}
Gaussian draws use the full covariance of the 27{,}501 unit training codes and are renormalized. A code is inside the support
if it passes PCA-based outlier detection \cite{jackson1991user} at the training codes' 95th percentile and a 50-nearest-neighbor
test \cite{sun2022out} at their 5th percentile. Novelty search \cite{lehman2011abandoning} ascends $\nu$ against the training
codes and an archive (rounds of 64 codes started at random training codes, 200 Adam steps at learning rate 0.01), with a hinge
penalty for leaving the support, and farthest-point selection \cite{gonzalez1985clustering} keeps a diverse subset. Training
codes have a median $\nu$ of $0.32$; the codes of Fig.~5 reach $0.65$ to $0.69$. For Fig.~\ref{fig:supp_samples}, the
neighbor test uses the median, and at most the median share of energy may lie in bands 8 to 15: the looser support drifts
into fast bands that render as aimless arm motion.

\paragraph{Motion grids.}
Both grids use the same six characters and scenes: a man in a parking garage, a young woman on a pier, a chrome robot in a white corridor, a knitted doll on a bedroom rug, a green cartoon ogre in a forest and an elderly woman in an autumn park. The simple motions are wave, arms up, squats, jumping jacks, spin, side steps, walk forward, kicks, bow, punches, hop and side bend. The two-step motions are arm then knee, knee then arm, lunge then jump, jump then lunge, squat then reach up, kick then turn around, step-clap twice then sway, two punches then knee, toe touch then side lean, kneel then stand and wave, slow arms then fast hops, and jumping jacks then toe touch. Clips come from the official Wan2.2-T2V-A14B sampler at its defaults (40 steps), at
$832 \times 480$, 81 frames at 16 fps, and all characters of a motion share a seed. Claude (Anthropic) wrote the prompts, each a fixed description of the character and scene followed by the motion's script. A full prompt reads: \emph{``A middle-aged man with short dark hair and a beard,
wearing a grey hoodie, black jeans and white sneakers, in an empty concrete parking garage with yellow lines on the floor. Static camera
at eye level, wide shot, the whole body visible from head to feet in the centre of the frame, no camera movement, no cuts, no other
people. The man starts standing still, facing the camera with the arms relaxed at the sides, then raises the right arm straight up
above the head and lowers it again, then lifts the left knee up to hip height and puts the foot back down, and ends standing still
facing the camera. Natural, physically plausible movement in one continuous take, 5 seconds.''}

\subsection{Baseline Comparison}
\paragraph{Baselines.}
Baselines run released code with default settings. DisMo \cite{ressler2025dismo} runs unquantized (CogVideoX-5B,
25 frames at $720 \times 480$, 50 steps, text guidance 6) and reads 25 frames sampled over the source. FlexiAct
\cite{zhang2025flexiact} tunes its motion embedding for 2{,}000 steps on 12 crops of the source made by its own rule, then generates
from the motion caption with the subject swapped for the target's, using its RefAdapter (weight 0.7), guidance 6 and 50 steps.
DisMo, FlexiAct and our CogVideoX version receive each target padded to $720 \times 480$ with its border color, never cropped; the padding is
removed before scoring. The text baseline uses Wan's official sampler (UniPC, 50 steps, guidance 5, 61 frames) with the prompt a concatenation of
``\emph{appearance}. \emph{motion caption}''.

\paragraph{Motion fidelity.}
\label{sec:supp_metric}
As published, the motion fidelity of Yatim \etal \cite{yatim2024space} compares directions only, so tracker jitter lets a still copy
of the target image score $0.49$, and it credits a camera move copied from the source. We fix these two points and keep the rest of the
score unchanged. CoTracker \cite{karaev2024cotracker} tracks a $55 \times 55$
grid over 25 frames, and camera motion is removed by chaining similarity transforms fitted to all points with RANSAC between
consecutive frames. The direction score on these tracks is multiplied by $\min(1, m_g / m_s)$, where $m$ is the mean path length, in
subject heights, of the subject's moving points (BiRefNet \cite{zheng2024birefnet} mask in the first frame; points that move more
than $5\%$ of the subject's height) in the generated and the source video; a still video scores $0$. The score is also sensitive to timing: shifting a motion
by $1/24$ of the clip removes $82\%$ of the margin between a perfect copy and a still video. An in-between video of an interpolation with weight $w$ is
scored $(1-w)$ times its fidelity to $A$ plus $w$ times to $B$. Its gate uses $(1-w)\,m_A + w\,m_B$ as the source's amount.

\paragraph{Number of frames.}
DisMo generates one output per frame its extractor reads, so on the 25-frame grid of Table~2 each of its frames lines
up with the source frame it came from, which a frame-to-frame motion score rewards; our videos (61 frames on Wan, 49 on
Cog) and FlexiAct's (49) are subsampled to it. On 20 or 10 frames, every method scores higher, since each step covers a
larger movement, and the ranking is unchanged, but our lead over DisMo grows (Table~\ref{tab:frames}), so 25 frames is least
favorable for us.

\begin{table*}[t]
\centering
\footnotesize
\setlength{\tabcolsep}{5pt}
\begin{tabular}{@{}lccccccc@{}}
\toprule
frames scored & text & FlexiAct & DisMo & \modelname{} (Cog-5B) & \modelname{} (Wan-5B) & Cog $-$ DisMo [95\% CI] & Wan $-$ DisMo [95\% CI] \\
\midrule
25 (Table~2) & $.256$ & $.361$ & $.378$ & $.421$ & $.410$ & $+.042$ $[+.015, +.070]$ & $+.032$ $[-.0004, +.065]$ \\
20           & $.280$ & $.399$ & $.412$ & $.463$ & $.452$ & $+.051$ $[+.021, +.081]$ & $+.040$ $[+.004, +.074]$ \\
10           & $.383$ & $.512$ & $.552$ & $.616$ & $.610$ & $+.064$ $[+.020, +.109]$ & $+.058$ $[+.010, +.107]$ \\
\bottomrule
\end{tabular}
\caption{\textbf{Motion fidelity vs. number of frames scored}, on 120 transfer pairs of Table~2. CI: 10{,}000
bootstrap resamples of the pairs.}
\label{tab:frames}
\end{table*}

\paragraph{Very different targets.}
The metric rewards reproducing the source's point trajectories, so it gives little credit to a body with very different
proportions that performs the motion along other paths, such as the desk lamp that lowers its head where the source lowers its
arms (Fig.~6), and full credit to a result that copies the source's limb paths, even by reshaping the target toward the
source's body as DisMo does in Fig.~4. Score and users therefore disagree on such targets (Fig.~\ref{fig:supp_shape}). Of the 300 further pairs of Sec.~4.3, on the 188 with a
non-human target (an animal or object; humanoid figures such as dolls count as humanoid), \modelname{} and DisMo have about
the same motion fidelity ($-0.019$, CI $[-0.049, 0.011]$), yet \modelname{} keeps the target better (image similarity
$+0.008$, CI $[0.003, 0.013]$; $+0.015$, CI $[0.008, 0.022]$, on the 81 non-human pairs of the 120), and users rank it first
in $76\%$ (top two in $97\%$) of their answers on non-human targets, against $48\%$ (top two in $93\%$) on humanoid ones.

\paragraph{User studies.}
Both studies show our CogVideoX version and all results at 25 frames, on web pages with questions
shuffled per participant and methods in random positions (Fig.~\ref{fig:supp_userstudy}). For transfer, we randomly draw five source videos
of Table~2 and two target images each, skipping targets that cannot perform articulated motion (e.g., a windmill). A question shows
the source, the target and the four results; participants pick the best and second-best on (i) how closely the source motion is
reproduced, (ii) whether it is still the target subject, taking nothing from the source (appearance, body shape, material or
background), and (iii) whether it looks natural and free of artifacts. For interpolation, each of the randomly selected ten pairs gets one random
target, and a question shows $A$, $B$, the target and each method's videos at $w = 0.25, 0.5, 0.75$, asking which interpolates
better while applying the motions properly to the target. Ours was chosen in $81\%$ of answers on the five pairs whose motions can
be done together (e.g., walking and raising the arms) and $71\%$ on the five where the blend must move from one to the other.

\paragraph{DisMo interpolation.}
Both source clips go through DisMo's motion extractor as 25 frames sampled over the whole clip, its input length, so the two
embedding streams align frame by frame. They are mixed with weights $1-w$ and $w$ and rescaled to the interpolated
root-mean-square magnitude, so that the blend does not shrink, before DisMo's own mapping; its end points ($w = 0, 1$) are plain
transfers of $A$ and $B$.

\FloatBarrier
\section{Additional Ablations}
\label{sec:supp_abl}

\paragraph{Temporal compression arms.}
Each arm of Table~4 trains its encoder and then its adapter for 30k steps on all three datasets. \emph{Pooled frames} replaces
the 16 cosines with 16 box functions, and its adapter holds each slot over one sixteenth of the output. \emph{Learned compression}
reads the frames averaged in the same 16 windows, so that its transformer sees as many tokens as the DCT, and 16 learned queries
pool over all of them; its adapter reads the code through learned frame queries in place of the inverse DCT. The code statistics
use the unit codes of the same 4{,}000 training clips for every arm. A shaken query has each frame shifted at random by up to
$4\%$ of its size; a vibrating one is shifted by $A \sin(2\pi i / P + \phi)$ in $x$ and $y$ at frame $i$, with $A = 3\%$ of the
frame and $P = 3$ or 5 frames (R@1 averaged over both periods and both grids); both are retrieved among the clean clips of the
other characters.
\vspace{-5pt}
\paragraph{Fig.~2a.}
The principal components are fitted on half of the synthetic motions; both bases are applied, without training, to the
encoder's own input (frozen DINOv2 tokens of the motion crop, displacement from the first frame).
\vspace{-5pt}

\paragraph{Without a time axis.}
An adapter receiving the same code at every frame scores 0.17 on 40 transfers, versus 0.33 for its matched DCT adapter (both trained for 20k steps on identical codes), and the subject barely moves.
\vspace{-5pt}

\addtocounter{table}{1}

{
\let\origtable\table\renewcommand{\table}[1][]{\origtable[!h]}
\begin{table}[]
\centering
\footnotesize
\setlength{\tabcolsep}{1pt}
\setlength{\aboverulesep}{1pt}\setlength{\belowrulesep}{1pt}\renewcommand{\arraystretch}{1.0}\setlength{\abovecaptionskip}{2pt}%
\begin{tabular*}{\columnwidth}{@{\extracolsep{\fill}}lccccc}
\toprule
encoder & R@1 synth\,$\uparrow$ & R@1 grid\,$\uparrow$ & mAP\,$\uparrow$ & mAP fast\,$\uparrow$ & same\,$\downarrow$ \\
\midrule
ours (full)                        & $\mathbf{100\%}$ & $\mathbf{75\%}$ & $.585$ & $\mathbf{.455}$ & $\mathbf{56\%}$ \\
w/o per-freq.\ balance             & $99\%$ & $74\%$ & $\mathbf{.592}$ & $.443$ & $62\%$ \\
w/o frame-0 sub        & $\mathbf{100\%}$ & $\mathbf{75\%}$ & $.575$ & $.435$ & $64\%$ \\
w/o $\mathcal{L}_{\text{nce}}$ & $83\%$ & $66\%$ & $.519$ & $.382$ & $60\%$ \\
RGB patches      & $82\%$ & $42\%$ & $.384$ & $.362$ & $59\%$ \\
\bottomrule
\end{tabular*}
\caption{\textbf{Encoder design}, each row retrained for fewer steps than the encoder of Table~1. Without the balance, overall mAP rises slightly because most motions are slow, while fast motions and same video get worse. Grid: simple motion grid; fast: its third with most high-frequency energy; same: same video (OOD).}
\label{tab:separation}
\end{table}
}

\FloatBarrier
\section{Additional Results}
\label{sec:supp_res}

\begin{figure*}[t]
  \centering
  \includegraphics[width=\textwidth]{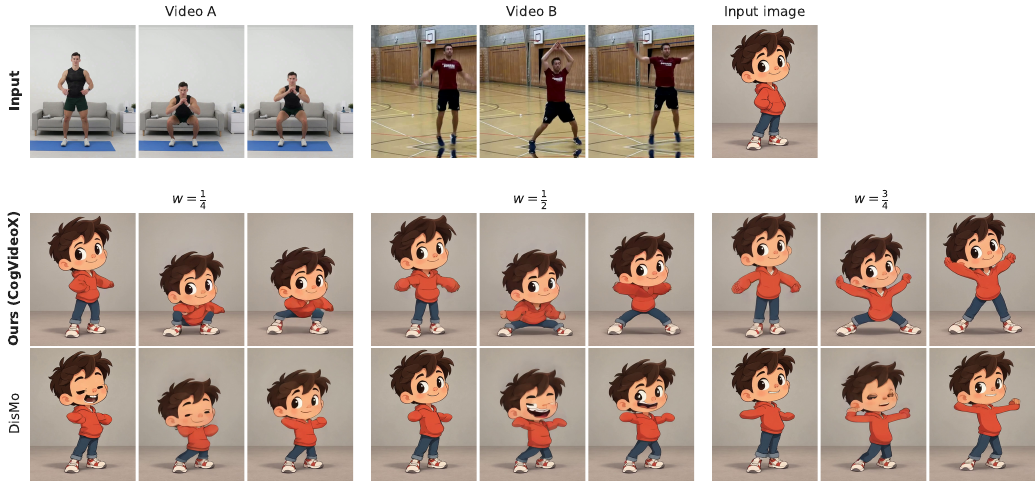}
  \caption{\textbf{Interpolation.} The codes of video A, a squat, and video B,
  a jumping jack, are blended with weight $w$ and applied to the same image;
  three moments of each output. Ours is our CogVideoX-5B version, on DisMo's base model. As $w$ grows, ours moves from a deep
  squat, through a wide squat with the arms out, to a jumping jack. DisMo, given
  the same blends of its per-frame embeddings, struggles with cleanly incorporating both motions at any $w$.}
  \label{fig:qual_interp}
\end{figure*}

\begin{figure*}[t]
  \centering
  \includegraphics[width=\textwidth]{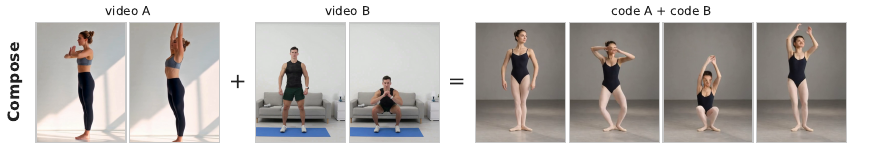}
  \caption{\textbf{Composition.} Adding the codes of an arm raise and a squat gives a squat with the arms overhead.}
  \label{fig:qual_compose}
\end{figure*}

\paragraph{Interpolation and composition.}
Fig.~\ref{fig:qual_interp} interpolates and Fig.~\ref{fig:qual_compose} composes two motions on one image (Sec.~3.3).

\begin{figure*}[t]
  \centering
  \includegraphics[width=\textwidth]{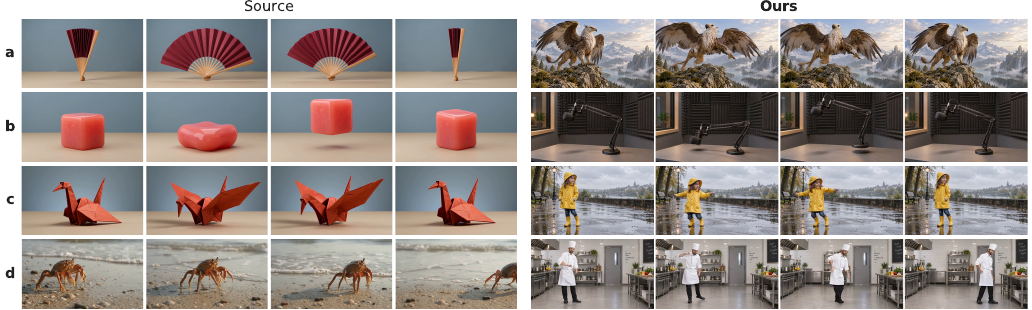}
  \caption{\textbf{Non-body motion.} (a)~A fan opening $\to$ a griffin spreading its wings. (b)~A jelly cube hopping $\to$ a
  microphone dipping, then hopping off the desk. (c)~A paper crane flapping $\to$ a girl spreading her arms. (d)~A crab scuttling
  sideways $\to$ a chef side-stepping.}
  \label{fig:qual_nonbody}
\end{figure*}

\begin{figure*}[t]
  \centering
  \includegraphics[width=\textwidth]{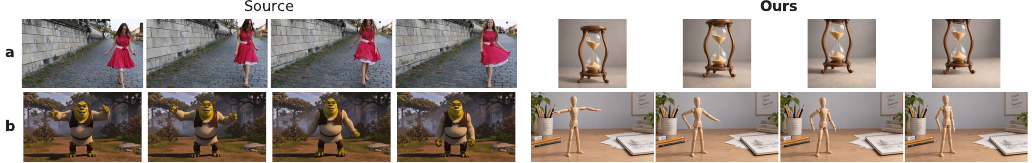}
  \caption{\textbf{Failure cases.} (a)~The source's camera motion carries over: the view of the hourglass moves with it.
  (b)~Fast motion is softened: the hops after the slow arm moves are smaller and slower than in the source.}
  \label{fig:supp_fail}
\end{figure*}

\begin{figure*}[t]
  \centering
  \includegraphics[width=\textwidth]{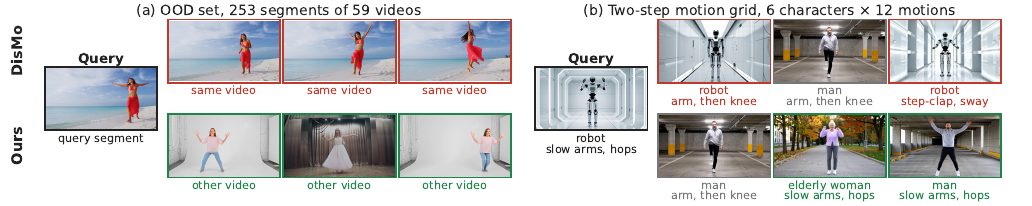}
  \caption{\textbf{What the nearest neighbors share.} The three nearest clips of a query in DisMo's space and in ours (Table~1's
  codes, cosine similarity), one frame per clip. Red: same video or character as the query; green: another video or character
  with the same motion.}
  \label{fig:supp_nn}
\end{figure*}

\begin{figure*}[t]
  \centering
  \includegraphics[width=\textwidth]{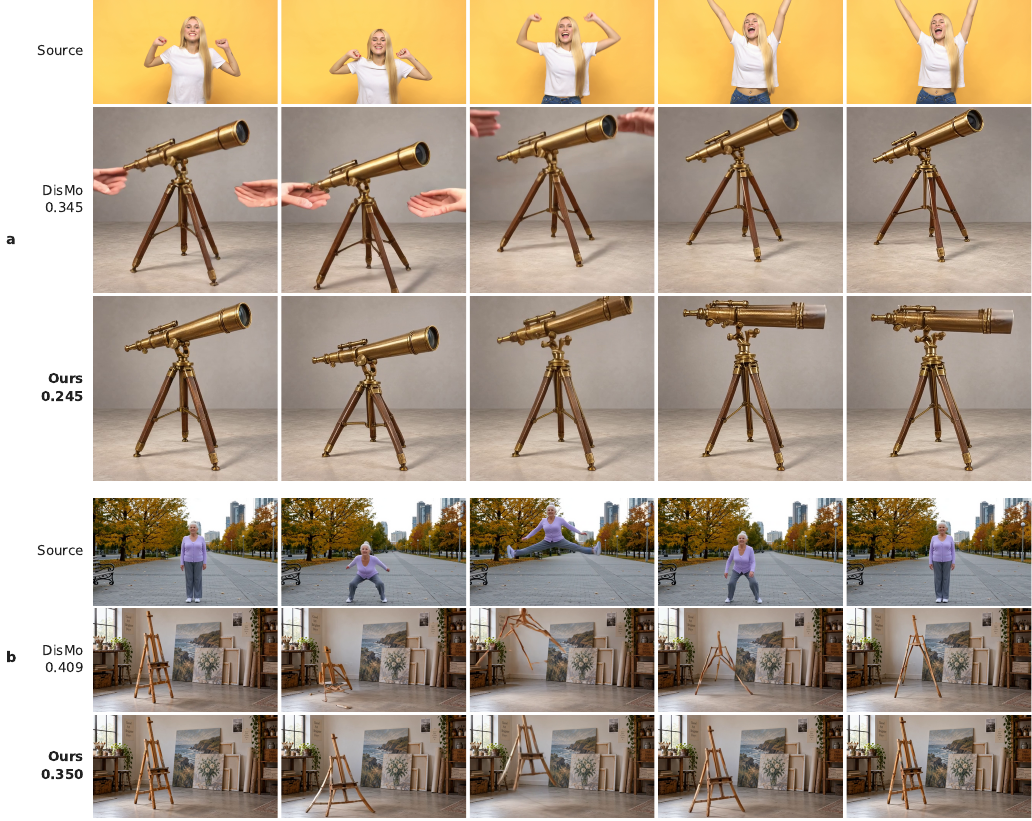}
  \caption{\textbf{Very different targets.} The number under each method is its motion fidelity: it favors DisMo, which copies the
  source's path at the target's expense, while ours (CogVideoX version) performs the motion in the target's own way. (a)~DisMo adds
  human hands to follow the arms; ours tilts its tube up as the arms go up (image similarity $0.911$ vs.\ $0.955$). (b)~DisMo's easel
  collapses and flies apart; ours splays its legs to go low and straightens again ($0.928$ vs.\ $0.964$).}
  \label{fig:supp_shape}
\end{figure*}

\paragraph{What the nearest neighbors share.}
Fig.~\ref{fig:supp_nn} shows the three nearest neighbors of one query from each test set. DisMo's neighbors tend to
share the query's appearance (its video or character) more often than ours do (same video and same character, Table~1).

\paragraph{The operations, measured.}
Table~\ref{tab:ops} is the full version of Table~3, on videos from our Wan2.2 model. Coarse to fine and intensity use 21 source
clips, each on 8 targets (people, humanoid figures, animals and objects); sampling uses humanoid targets. Four of the 16 bands
already carry $80\%$ of a motion without changing the target's appearance, and the re-encoded code moves along the intensity
direction in $96\%$ of pairs. A composed video lies closer to each motion's code than the other motion alone does ($+0.27$, CI
$[0.24, 0.29]$) and, on tracks, follows each source more ($+0.014$, CI $[0.011, 0.018]$, toward B; $+0.010$, CI $[0.007, 0.013]$,
toward A). Sampled codes keep the visual quality as well as real codes from source clips.

\begin{table*}[t]
\centering
\footnotesize
\setlength{\tabcolsep}{4pt}
\begin{tabular}{@{}lllc@{}}
\toprule
operation & setting & measure & values \\
\midrule
coarse to fine (168) & bands $k < 2$ / $4$ / $8$ / all $16$ & motion fidelity (tracks) & $.267$ / $.350$ / $.370$ / $.410$ \\
 & & share of the motion (tracks) & $45$ / $80$ / $95$ / $100\%$ \\
 & & image similarity & $.909$ / $.903$ / $.901$ / $.902$ \\
 & & source found among 21 (code) & $23$ / $55$ / $88$ / $91\%$ \\
\midrule
intensity (168) & $\alpha = -2$ / $-1$ / $0$ / $+1$ / $+2$ & subject motion, body heights (tracks) & $.05$ / $.26$ / $.76$ / $1.20$ / $1.38$ \\
 & & image similarity & $.944$ / $.932$ / $.902$ / $.860$ / $.826$ \\
 & & re-encoded code along $d$ (code) & $-.71$ / $-.54$ / $-.31$ / $-.09$ / $+.07$ \\
\midrule
composition (105) & A + B / A alone / B alone & fidelity to A (tracks) & $.485$ / $.503$ / $.475$ \\
 & & fidelity to B (tracks) & $.493$ / $.479$ / $.505$ \\
 & & cosine to A's code (code) & $.34$ / $.49$ / $.06$ \\
 & & cosine to B's code (code) & $.33$ / $.06$ / $.49$ \\
\midrule
sampling & Gaussian (150) / novelty (60) / transfers (109) & temporal consistency & $.978$ / $.979$ / $.969$ \\
 & & image similarity & $.896$ / $.898$ / $.882$ \\
 & & near-still videos & $14$ / $13$ / $8\%$ \\
\bottomrule
\end{tabular}
\caption{\textbf{The operations, measured.} Tracks: the measures of Sec.~\ref{sec:supp_metric} (fidelity: direction score on camera-removed tracks, times
the amount gate where a source exists); code: the video re-encoded by the model's encoder, compared with the source codes (chance
of finding the source: $5\%$). Image similarity: to the target image.}
\label{tab:ops}
\end{table*}

\begin{figure*}[t]
  \centering
  \includegraphics[width=\textwidth]{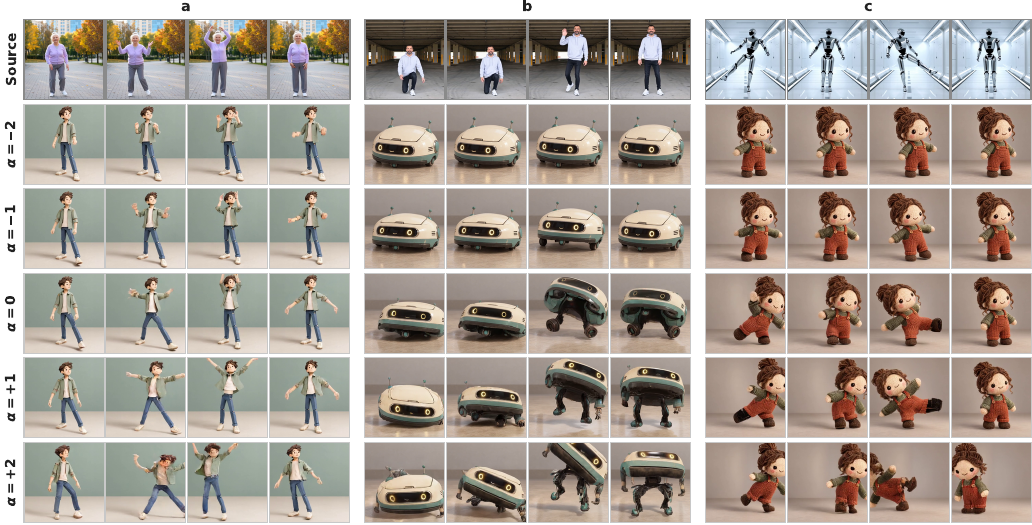}
  \caption{\textbf{Intensity.} Each code is moved along the intensity direction, $\hat c + \alpha d$, from $\alpha = -2$ to $+2$
  ($\alpha = 0$ is the plain transfer). The subject's motion on point tracks grows at every step: from $0.04$ to $2.44$, $0.00$ to
  $1.77$ and $0.07$ to $2.42$ body heights in (a)--(c).}
  \label{fig:supp_intensity}
\end{figure*}

\begin{figure*}[t]
  \centering
  \includegraphics[width=\textwidth]{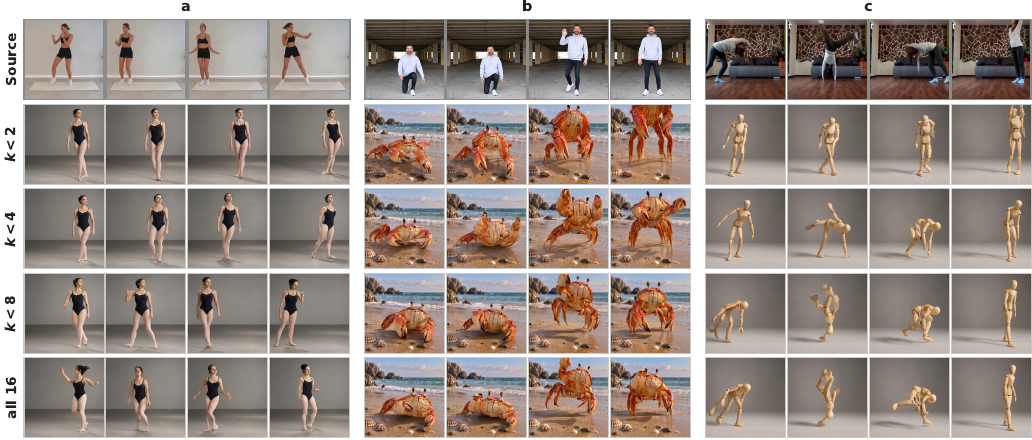}
  \caption{\textbf{Coarse to fine.} Each code keeps only its bands $k < 2$, $4$ or $8$, or all 16. With two bands the target follows at most
  the slow course of the motion; arm moves (a), the kneel (b) and the cartwheel (c) appear as bands are added.}
  \label{fig:supp_coarse}
\end{figure*}

\begin{figure*}[t]
  \centering
  \includegraphics[width=\textwidth]{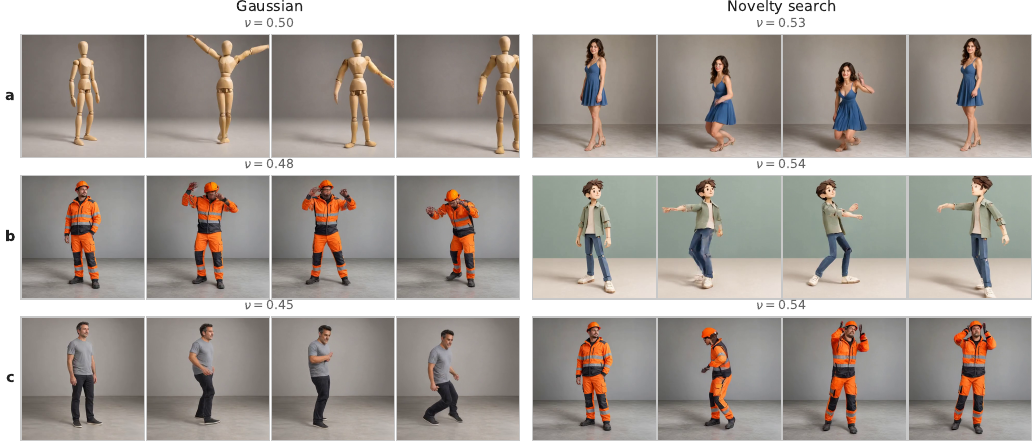}
  \caption{\textbf{Motions with no source video.} Codes drawn from the Gaussian over the training codes (left) and found by novelty
  search (right), each rendered on a target image. $\nu$: novelty, one minus the mean cosine similarity of the code to its three
  nearest training codes (training median $0.32$).}
  \label{fig:supp_samples}
\end{figure*}

\paragraph{Further tests on latent space.}
Fig.~\ref{fig:supp_intensity} scales
three motions along the intensity direction, Fig.~\ref{fig:supp_coarse} cuts three codes to fewer bands, and Fig.~\ref{fig:supp_samples}
renders codes that no clip performed, drawn from the Gaussian or found by novelty search. All use the Wan2.2 model at 50 steps.

\paragraph{Figure~4 on CogVideoX.}
Fig.~\ref{fig:supp_adapt_cog} adds to Fig.~4 with our CogVideoX-5B version.

\begin{figure*}[t]
  \centering
  \leavevmode\input{figures/qual_adapt_cog_anim}
  \caption{\textbf{Figure~4 with ours on CogVideoX-5B.} The examples of Fig.~4, with the CogVideoX-5B version of our method
  below the dotted line, on the model FlexiAct and DisMo build on. It also performs each motion with the target's own body: the
  mannequin twirls, and the doll crouches and raises its arms without taking the golem's shape. Animates in Adobe Acrobat.}
  \label{fig:supp_adapt_cog}
\end{figure*}

\paragraph{Limitations.}
The encoder and adapters train on body motion, but we find that it can generalize also (Fig.~\ref{fig:qual_nonbody}).

The code is nearly blind to timing finer than one latent frame. Starting a test clip half a latent frame later changes its code
(one minus cosine similarity) by $0.17$, no more than encoding the same clip from a crop instead
of the full frame ($0.20$, 120 clips); a full latent frame shift gives $0.40$, and a different clip $0.89$. Separating camera motion from body motion
is outside the scope of this paper, so strong camera motion in the source can carry over into the result. Fig.~\ref{fig:supp_fail}
shows both failure cases: camera motion carried over and softened fast motion.

\begin{figure*}[t]
  \centering
  \includegraphics[width=\textwidth]{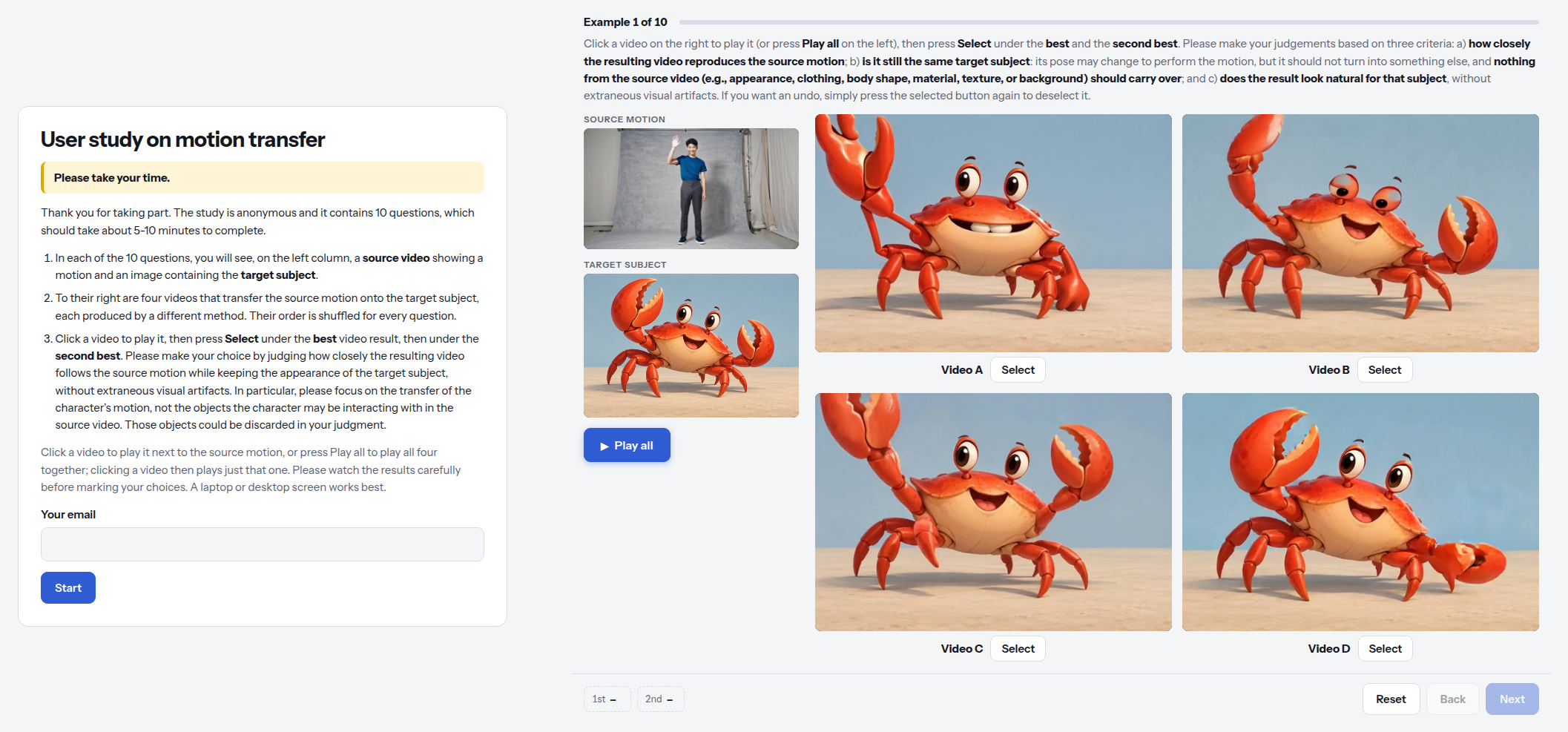}\\[1pt]
  {\small (a) Transfer study}\\[6pt]
  \includegraphics[width=\textwidth]{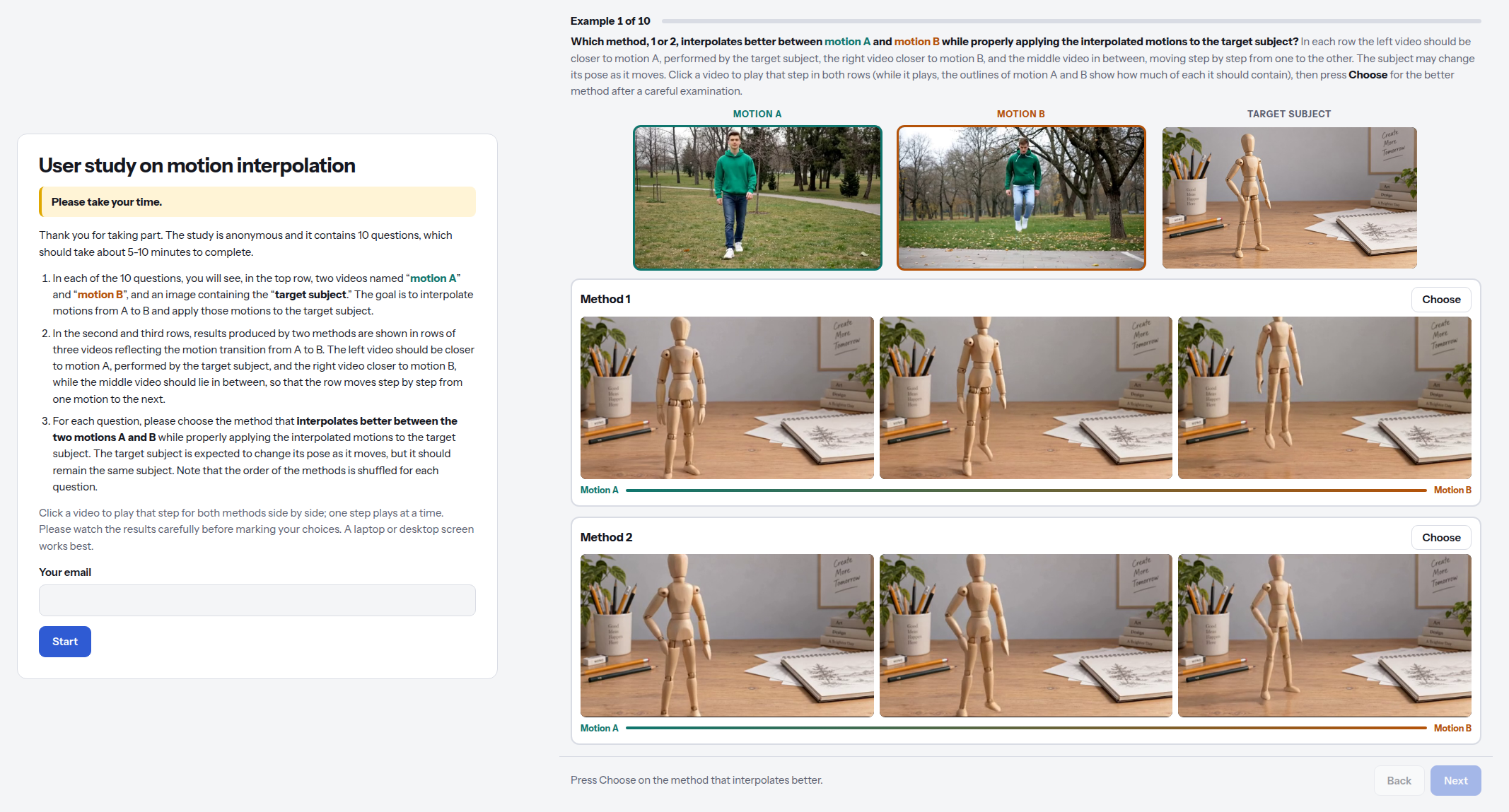}\\[1pt]
  {\small (b) Interpolation study}
  \caption{\textbf{User study pages.} Start page (left) and one question (right) of each study. (a)~Transfer: the source motion,
  the target subject and the four results in random order; participants select the best and the second best. (b)~Interpolation:
  motions $A$ and $B$, the target subject and each method's three in-between videos ($w = 0.25, 0.5, 0.75$) in random order;
  participants choose the better method. Emails were collected optionally, only for payment, as this was a paid study.}
  \label{fig:supp_userstudy}
\end{figure*}

\FloatBarrier

{
    \small
    \bibliographystyle{ieeenat_fullname}
    \bibliography{main}
}